\documentclass{article} % For LaTeX2e
\usepackage{gpl_iclr2022_conference,times}

\usepackage{amsmath,amsfonts,bm}

\def\eqref#1{equation~\ref{#1}}
\def\1{\bm{1}}

\DeclareMathAlphabet{\mathsfit}{\encodingdefault}{\sfdefault}{m}{sl}
\SetMathAlphabet{\mathsfit}{bold}{\encodingdefault}{\sfdefault}{bx}{n}

\usepackage{hyperref}
\usepackage{url}
\usepackage{graphicx} % lk

\newcommand{\tabincell}[2]{\begin{tabular}{@{}#1@{}}#2\end{tabular}} %lk

\title{An Empirical Study and Analysis of Learning Generalizable Manipulation Skill in the SAPIEN Simulator
}

\author{Kun Liu$^1$, Huiyuan Fu$^1$, Zheng Zhang$^1$, Huanpu Yin$^2$ \\
Beijing University of Posts and Telecommunications$^1$\\ 
Beijing Technology and Business University$^2$\\
\texttt{\{liu\_kun,fhy,zhangzheng\}@bupt.edu.cn,yinhuanpu@btbu.edu.cn} \\
}

\iclrfinalcopy
\begin{document}
\maketitle

\begin{abstract}
This paper provides a brief overview of our submission to the no interaction track of SAPIEN ManiSkill Challenge 2021.
Our approach follows an end-to-end
pipeline which mainly consists of two steps:
we first extract the point cloud features of multiple objects;
then we adopt these features to predict the action score of the robot simulators through a deep and wide transformer-based network.
More specially, %to give guidance for future work,
to open up avenues for exploitation of learning manipulation skill,
we present an empirical study that includes a bag of tricks and abortive attempts.
Finally, our method achieves a promising ranking on the leaderboard.
All code of our solution is available at  https://github.com/liu666666/bigfish\_codes.
\end{abstract}

\section{Introduction}

We perceive the real world by seeing the 3D environments and manipulating 3D objects surrounding us.
Nowadays, researchers in the computer vision and deep learning community pay more and more attention to establishing artificial intelligence that can recognize and interact with objects in 3D scenes as human beings do.
To this end, SAPIEN Open-Source Manipulation Skill Challenge, termed as ManiSkill Challenge \cite{mu2021maniskill}, was introduced to concentrate on learning unseen objects manipulation skills.

\begin{figure*}[t]
\begin{minipage}[b]{0.96\linewidth}
  \centering
  \centerline{\includegraphics[width=12cm,height=7.5cm]{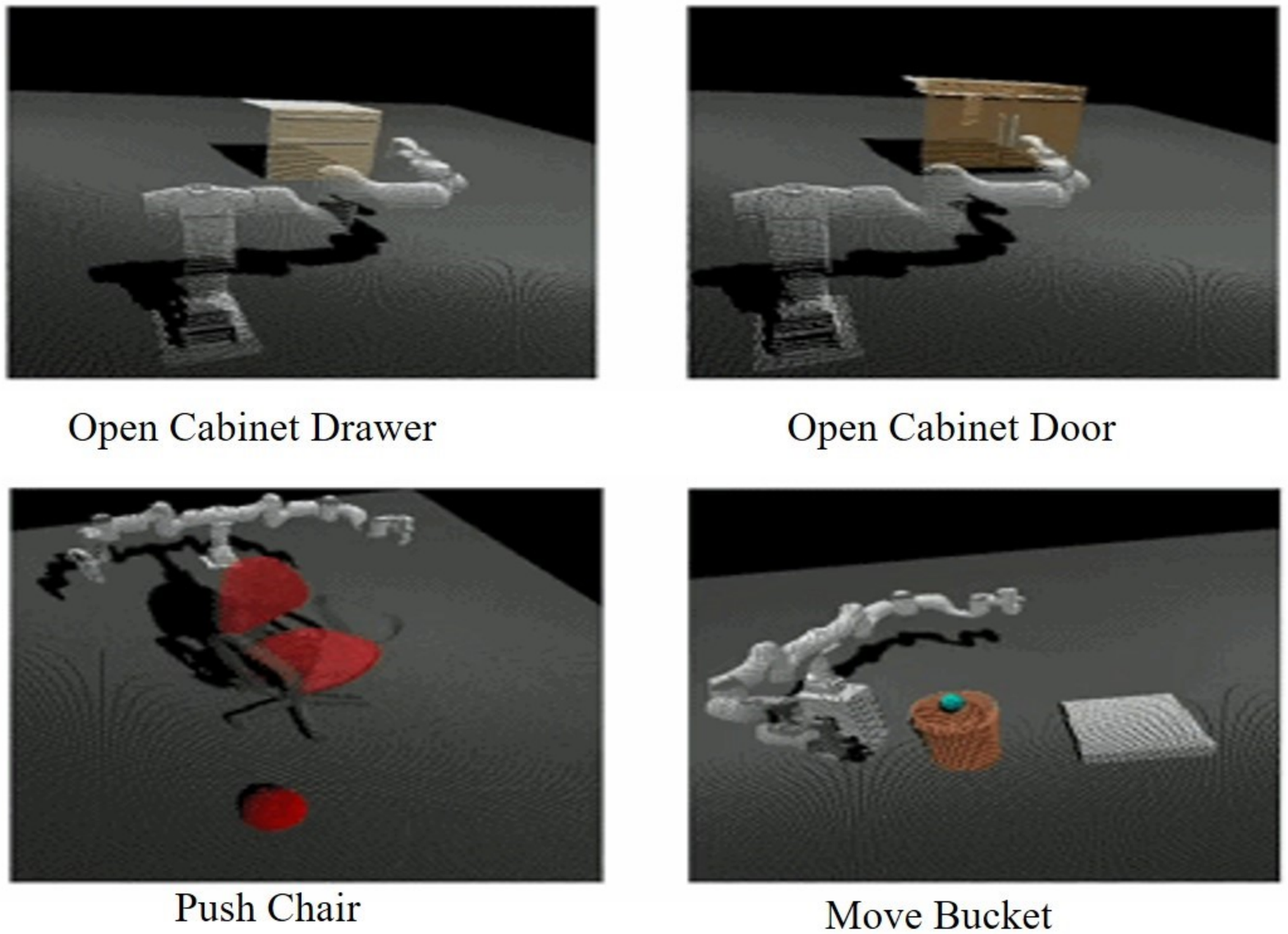}}
\end{minipage}
\caption{The samples of four manipulation tasks: Open Cabinet Door, Open Cabinet Drawer, Push Chair, and Move Bucket}
\label{fig:01}
\end{figure*}

Based on the SAPIEN Manipulation Skill Benchmark (ManiSkill Benchmark) \cite{xiang2020sapien}, ManiSkill challenge utilizes a variety of articulated objects in a full-physics simulator to accomplish 4 tasks:
Opening Cabinet Door, Opening Cabinet Drawer, Pushing Chair, and Moving Bucket.
Figure \ref{fig:01} demonstrates four tasks according to the examples from ManiSkill Benchmark, which consists of 162 objects and  ~36,000 successful demonstrations in total. It is noted that these tasks require manipulating the unseen test objects after learning on training objects where the training and testing samples are from the same class.

To learn the generalizable manipulation skill,
we propose an end-to-end framework that extracts the 3D visual features and predicts the action of the robot arms. More specially, we adopt the PointNet \cite{qi2017pointnet} to learn the point cloud features and then utilize a transformer \cite{vaswani2017attention} based network to output the action of robots.
Moreover, we provide an empirical study that is mainly composed of useful tricks and failure trials to learn generalizable manipulation skills.

\section{Related Work}
To clarify, we classified the related works into two aspects: 3D simulators and point cloud feature extractors.  % and transformer networks

At present, more and more 3D simulators are constructed for Embodied AI.
AI2-THOR \cite{kolve2017ai2} is a platform that provides the manipulation capabilities of agents, especially the interaction with actionable objects in four room categories: bed, bathroom, kitchen, and living room.
The Interactive Gibson Environment (iGibson) \cite{xia2020interactive} can render dynamical environments by using photogrammetry for their room construction.  Besides, iGibson provides users with ten fully functional and common robotic agents, which can improve the application in the real world.
This challenge adopts the SAPIEN environment that supplies the interaction with diverse articulated objects without any commercial software.
More details about 3D simulators can be found in this survey \cite{duan2022survey}.

This challenge simulates a moving panoramic camera that provides ego-centric point clouds or RGB-D images.
%Following \cite{mu2021maniskill}, we adopt the point clouds as the 3D visual input and utilize the PointNet \cite{qi2017pointnet} as the feature extractor.
Inspired by success brought by the deep learning techniques \cite{mu2021maniskill,liu2018t}, we adopt the point clouds as the 3D visual input and utilize the PointNet \cite{qi2017pointnet} as the feature extractor.
PointNet extracts the powerful point cloud feature by learning pointwise features independently through several MLP layers and modeling global features via a max-pooling layer.
PointNet ++ \cite{charles2017deep} improves PointNet by designing a hierarchical network
to model fine geometric architecture from the neighborhood of each point.
Furthermore, Structural Relational Network (SRN) was proposed in \cite{duan2019structural}
to learn structural relational information between different local units using MLP.
More details about point cloud can be referred to this paper \cite{guo2020deep}.

\section{Method}
In this section, we introduce our method briefly.
Basically, we adopt an end-to-end architecture to
extract the point cloud features and predict the action of robot arms. Specifically, we utilize the PointNet to extract the 3D feature of manipulated objects and develop a deep and wide transformer-based network to output the direction and the moving distance of the arm simulators.

\section{Experiments}
To give guidance for future work, we present an empirical study that includes a bag of tricks and abortive attempts.
Specially, we tried various combinations of training samples with different trajectories, different head numbers in transformer architecture, the depth of transformer network, batch size, iteration number, and so on.

\begin{table}[h]
\caption{The submission result on Opening Cabinet Drawer task}
\label{01}
\begin{center}
\begin{tabular}{cccc}
\tabincell{c}{Row} &\tabincell{c}{Training \\ Success Rate}  &\tabincell{c}{Testing \\ Success Rate} &\tabincell{c}{Detail}
\\ \hline \\
1   &0.37         &0.245   &6 Block, 4 Head, 150K Iteration, 128 Batch Size(baseline) \\
2   &0.47         &0.336  &Same as above besides 300K Iteration, 1024 Batch Size \\
3   &0.69         &0.404  &6 Block, 64 Head, 900K Iteration, 1024 Batch Size \\
4   &\bf{0.70}         &\bf{0.508}  &\bf{12 Block, 64 Head, 900K Iteration, 1024 Batch Size}  \\
5   &0.70         &0.496  &Just submit above model again  \\
6   &0.68         &0.444  &16 Block, 16 Head, 900K Iteration, 1024 Batch Size \\
\end{tabular}
\end{center}
\end{table}

\begin{table}[h]
\caption{The submission result on Opening Cabinet Door task}
\label{02}
\begin{center}
\begin{tabular}{cccc}
\tabincell{c}{Row} &\tabincell{c}{Training \\ Success Rate}  &\tabincell{c}{Testing \\ Success Rate} &\tabincell{c}{Detail}
\\ \hline \\
1   &0.30         &0.205  &6 Block, 4 Head, 150K Iteration, 128 Batch Size(baseline) \\
2   &0.34         &0.244  &12 Block, 16 Head, 300K Iteration \\
3   &0.53         &0.384  &12 Block, 16 Head, 900K Iteration  \\
4   &0.56         &0.320  &Same as above but trained with various trajectories  \\
5   &0.52         &0.280  &12 Block, 64 Head, 900K Iteration  \\
6   &0.58         &0.388  &16 Block, 16 Head, 900K Iteration  \\
7   &\bf{0.60}         &\bf{0.416}  &\bf{Fine-tuned on above best model} \\
\end{tabular}
\end{center}
\end{table}

We report the results of four tasks on Table \ref{01}, Table \ref{02}, Table \ref{03} and Table \ref{04}, respectively.
We can observe that the opening cabinet drawer and opening cabinet door is easier than moving a bucket and pushing the chair. Specially, the former tasks achieve more than 0.4 success rate while the latter tasks obtain less than 0.3 success rate.
To our surprise,  models that perform close on the training set can differ by more than 10\% points on the testing set, such as row 3 and row 4 in Table \ref{01}.
Besides, as shown in row 4 and row 5 in Table \ref{01}, submitting the same model can achieve a 1.2\% increment.
Next, we summary up some advice from two aspects: useful tricks and failed attempts.

In this section, we list a bag of useful tricks.
\begin{enumerate}
\item Compared with baseline \cite{mu2021maniskill}, the model trained with a larger batch size and longer iteration number can bring significant performance improvement.
Taking row 1 and row 2 in Table \ref{01} as an example, larger batch size and longer iteration number results in 9.1\% absolute increment from 24.5\% to 33.6\% on the Opening Cabinet Drawer task.
Consequently, we adopt these hyper-parameters settings for the following experiments.
\item Compared with the model trained with suitable iteration and batch size, to some degree, deeper (more blocks) and wider (more heads) transformer networks increase the success rate drastically. Taking row 2 and row 3 in Table \ref{01} as an example,
wider networks, e.g., from 4 heads to 64 heads, bring 6.8\% improvement in terms of testing success rate. As for deeper networks, Taking row 3 and row 4 in Table \ref{01} as an example, 12 blocks outperform 6 blocks significantly ( 10.4\% ) on the testing set.
\end{enumerate}

\begin{table}[h]
\caption{The submission result on Moving Bucket task}
\label{03}
\begin{center}
\begin{tabular}{cccc}
\tabincell{c}{Row} &\tabincell{c}{Training \\ Success Rate}  &\tabincell{c}{Testing \\ Success Rate} &\tabincell{c}{Detail}
\\ \hline \\
1   &0.15         &0.115   &6 Block, 4 Head, 150K Iteration, 128 Batch Size(baseline) \\
2   &0.24         &0.280   &6 Block, 16 Head, 900K Iteration, 1024 Batch Size \\
3   &0.30         &0.292   &12 Block, 16 Head, 900K Iteration, 1024 Batch Size \\
4   &0.31         &0.280   &6 Block, 64 Head, 900K Iteration, 1024 Batch Size  \\
5   &0.35         &\bf{0.372}  &\bf{12 Block, 64 Head, 900K Iteration, 1024 Batch Size}   \\
6   &\bf{0.39}         &0.288  &16 Block, 16 Head, 900K Iteration, 1024 Batch Size \\
\end{tabular}
\end{center}
\end{table}

On the contrary, we present some useless efforts.
\begin{enumerate}
\item We construct different training sets by generating samples using various trajectory lengths, such as 100 trajectories, 200 trajectories, and 300 trajectories. Then we combine them as several training sets to learn skills. As shown in row 4 of Table \ref{02}, we achieve improvement on the training set but not on testing set.
\item We also tried some different transformer variants, such as \cite{bhojanapalli2020low}. Regrettably, as shown in row 6 of Table \ref{04}, no improvement is achieved.
\item We believe that approaches that perform poorly on the training set will also not achieve good results on the testing set. Hence, we did not submit some solutions to the challenge since they do not get promising results on the training set. This kind of solution includes replacing 64 heads with 256 heads, replacing 12 blocks with 18 blocks, and different learning rates.
\end{enumerate}

\begin{table}[h]
\caption{The submission result on Pushing Chair task}
\label{04}
\begin{center}
\begin{tabular}{cccc}
\tabincell{c}{Row} &\tabincell{c}{Training \\ Success Rate}  &\tabincell{c}{Testing \\ Success Rate} &\tabincell{c}{Detail}
\\ \hline \\
1   &0.18         &0.130   &6 Block, 4 Head, 150K Iteration, 128 Batch Size(baseline) \\
2   &0.26         &\bf{0.232}   &\bf{12 Block, 16 Head, 900K Iteration, 1024 Batch Size} \\
3   &0.30         &0.224   &6 Block, 64 Head, 900K Iteration, 1024 Batch Size \\
4   &\bf{0.32}         &\bf{0.232}   &\bf{12 Block, 64 Head, 900K Iteration, 1024 Batch Size} \\
5   &0.26         &0.164   &16 Block, 16 Head, 900K Iteration, 1024 Batch Size   \\
6   &0.31         &0.224   &Adopt a transformer variant \cite{bhojanapalli2020low} \\
\end{tabular}
\end{center}
\end{table}

\begin{table}[h]
\caption{The final result and ranking of all participating teams}
\label{05}
\begin{center}
\begin{tabular}{ccccccc}
\tabincell{c}{Rank}
&\tabincell{c}{Team \\ Name}
&\tabincell{c}{Final \\ Score}
&\tabincell{c}{Moving \\ Bucket}
&\tabincell{c}{Opening \\ Door}
&\tabincell{c}{Opening \\ Drawer}
&\tabincell{c}{Pushing \\ Chair}
\\ \hline \\
1   &Silver-Bullet-3D &0.5740         &0.602  &0.552  &0.744  &0.398  \\
2   &Fattonny         &0.4070         &0.328	&0.462	&0.512	&0.326 \\
3   &bigfish(ours) &0.3435         &0.310	&0.394	&0.466	&0.204 \\
4   &MI &0.3320         &0.212	&0.412	&0.454	&0.25 \\
5   &SieRra11799 &0.1890         &0.170	&0.122	&0.300	&0.166  \\
6   &ic &0.1880         &0.084	&0.176	&0.356	&0.136  \\
7   &zhihao &0.1835         &0.140	&0.170	&0.268	&0.156  \\
\end{tabular}
\end{center}
\end{table}

Table \ref{05} shows the final result and ranking of all participating teams.
Finally, we got a top-3 ranking in this no interaction track of the ManiSkill challenge.
More importantly, to promote the research on manipulating unseen objects in the SAPIEN simulator, we release our models and scripts on Github as a new baseline.

\section{Conclusion}
In this technical report, we present a brief description of our solution to the no interaction track of SAPIEN ManiSkill Challenge 2021.
Our method utilizes an end-to-end
pipeline which mainly consists of two steps:
first, we extract the point cloud features of multiple objects;
then we adopt these features to predict the action score of the robot simulators via a transformer-based network.
More specially, to give guidance for future work,
we provide an empirical study that includes a bag of tricks and some useless attempts.
Finally, our method achieves a top ranking on the leaderboard.
In future works, we will further explore to
manipulate unseen objects through designing suitable learning from demonstrations (LfD) algorithms.

\section{Acknowledgement}
This work is supported in part by National Key R\&D Program of China (2019YFB2102202).

\bibliography{gpl_iclr2022_conference}
\bibliographystyle{gpl_iclr2022_conference}

\end{document}